\documentclass[11pt,a4paper]{article}
\usepackage[margin=25mm]{geometry}
\usepackage[T1]{fontenc}
\usepackage{lmodern}
\usepackage{amsmath,amsfonts}
\usepackage{graphicx,booktabs,array}
\usepackage{subcaption,placeins,enumitem}
\setlist{nosep}
\usepackage[american]{babel}
\usepackage{csquotes}
\usepackage[backend=biber,style=numeric,sorting=none,natbib=true]{biblatex}
\usepackage{hyperref}
\hypersetup{hidelinks}
\author{Furkan Gursoy}
\date{}
\title{CartoVec: General-Purpose Location Embeddings\\from Open Map Data}
\begin{document}
\maketitle
\begin{abstract}

Geographic coordinates are common inputs in applied machine learning, but latitude and longitude describe position rather than the surrounding environment. Image-grounded location encoders learn this context from satellite or ground-level imagery; structured open map data offers another source of supervision whose utility for continuous coordinate encoders remains less explored. We introduce CartoVec, a coordinate encoder trained to predict Overture-derived land cover, land use, transportation, place and connectivity targets at multiple spatial scales, without image inputs or downstream-task labels. We evaluate frozen embeddings on Census outcomes, Airbnb prices and Philadelphia crime in the US Northeast using spatial block cross-validation. CartoVec has the highest mean performance on most Census outcomes and the highest average across the full suite, with several comparisons close across folds. The findings support open map data as a practical source of reusable location representations with a coordinate-to-vector interface that requires no map lookup at inference. The open-weight model is available at \href{https://huggingface.co/gursoyfurkan/CartoVec}{huggingface.co/gursoyfurkan/CartoVec}.

\end{abstract}


\section{Introduction}\label{introduction}

Many prediction problems depend on the environment around a location. Tasks such as predicting property values, forecasting local service demand, estimating last-mile delivery times and assessing site suitability can benefit from information about road access, land use, local amenities and the built environment. Coordinates often accompany the records used for such tasks, but obtaining useful environmental features often involves collecting relevant geographic data, interpreting its attributes and designing task-specific features.

Raw coordinates provide a limited starting point. A linear model given latitude and longitude can represent only a planar trend. Decision trees can capture sharper variation by partitioning coordinate space, while positional encodings provide geometric features through nonlinear transformations of coordinates. These approaches can be effective. Without pretraining on environmental data, however, the downstream model must learn useful spatial patterns from the labels available for each task.

Location encoders offer an alternative: a coordinate is mapped to a vector that can be reused across tasks. The encoder is trained separately from downstream models, which are then fitted on its fixed output vectors. This frozen-encoder approach allows the same learned location features to support different outcomes without retraining the encoder.

Several learned encoders obtain environmental context by aligning coordinates with imagery. SatCLIP \citep{satclip} uses satellite images, and GeoCLIP \citep{geoclip} uses geotagged photographs. Their training requires image collection and processing, although their released coordinate encoders do not require images at inference. In contrast, untrained positional features, such as random Fourier features (RFF) \citep{rff}, transform coordinates into a form that makes spatial variation easier to model without learning what is present at those coordinates.

Open map data offers a different source of environmental supervision. Overture Maps \citep{overture} catalogues places, buildings, transportation networks and land characteristics as geographic records with locations, geometries and descriptive attributes. Information such as business categories, road connectivity and land-use boundaries is supplied as attributes or geometries, rather than features that a model must infer from pixels. Some of this information can also be inferred from imagery, but the two sources differ in representation and coverage. Using structured map attributes to supervise reusable, continuous coordinate encoders remains less explored.

\begin{figure}[htbp]\centering
\includegraphics[width=\linewidth,height=0.65\textheight,keepaspectratio]{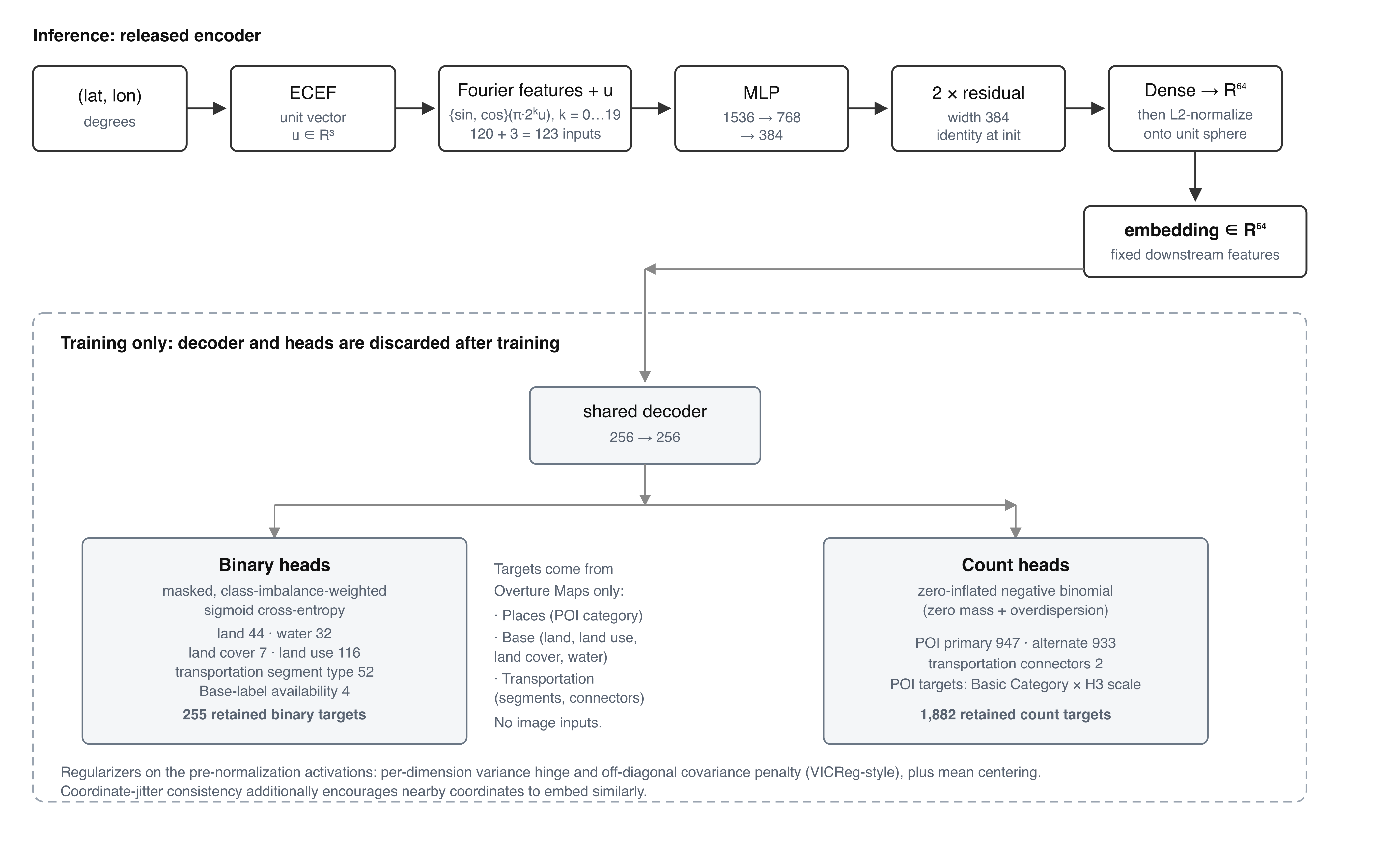}
\caption{CartoVec architecture. The inference path converts latitude and longitude in degrees to a spherical unit vector, adds Fourier positional features, and produces a unit-normalized 64-dimensional embedding through a multilayer perceptron with residual blocks. The dashed region contains the training-only decoder and prediction heads. Map-derived targets and regularization train the encoder; the decoder and heads are discarded at inference.}
\label{fig:1}
\end{figure}

This paper asks whether structured open map data can train a continuous coordinate encoder competitive with image-grounded encoders. We develop \textbf{CartoVec}, a 64-dimensional coordinate encoder supervised by multi-task prediction of Overture-derived targets, with no imagery supplied to the training pipeline. We train and evaluate the model in the US Northeast, a study area chosen to keep map processing and training computationally manageable. Frozen embeddings are evaluated under spatial cross-validation. Figure~\ref{fig:1} shows CartoVec's architecture and its coordinate-only inference path.

\textbf{Contributions.}
\begin{enumerate}
\item \textbf{A reusable, open-weight location encoder.} We develop CartoVec, a 64-dimensional encoder learned from structured open map attributes, and release its weights for reuse as fixed location features in downstream models. The released model covers the US Northeast.
\item \textbf{A shared evaluation across downstream tasks.} We evaluate CartoVec and nine baselines under spatial block cross-validation on eight prediction tasks: six Census outcomes, Airbnb prices and Philadelphia crime counts. CartoVec has the highest mean score on most Census targets and the highest macro-average across the full suite. Comparisons across five prediction-network architectures assess how probe choice affects these results.
\end{enumerate}

\FloatBarrier
\section{Related Work}\label{related-work}

\textbf{Open map data.} OpenStreetMap \citep{osm-background} is a foundational open geographic dataset, built by volunteer mapping and covering places, buildings, roads, land use and land cover as geographic features with geometries and descriptive tags. Its scale and coverage make it a central source for open geographic information.

Overture Maps \citep{overture} consolidates contributions from several providers into standardized themes, including Places, Buildings, Transportation and Base. Its typed schema provides category fields suitable for prediction targets, and its published coarser place grouping lets us aggregate sparse category counts. OpenStreetMap is one of several sources contributing to Overture. We ingest Overture rather than raw OpenStreetMap and describe the supervision as Overture-derived throughout.

\textbf{Positional encodings.} Geometric location encodings transform coordinates into features for spatial prediction \citep{locationreview}. The simplest representations are computed rather than learned. Random Fourier features \citep{rff} expand a coordinate into a bank of sinusoids at different frequencies, allowing a linear prediction model to represent richer spatial variation. Space2Vec \citep{space2vec} and Sphere2Vec \citep{sphere2vec} provide multi-scale grid and spherical positional features, respectively. Our comparison includes pretrained encoders built on these features, alongside untrained random Fourier features.

\textbf{Learned encoders trained with imagery.} A second family learns a coordinate encoder by aligning it with visual data. SatCLIP \citep{satclip} aligns a spherical-harmonic location encoder with Sentinel-2 satellite imagery through contrastive learning. GeoCLIP \citep{geoclip} aligns a multi-scale positional encoder with geotagged photographs used for worldwide geolocalization. Both provide globally pretrained coordinate encoders. Related work includes GPS2Vec \citep{gps2vec} and contrastive spatial pretraining \citep{csp}.

\textbf{Map-derived representations.} Structured map data can support representations of place categories and geographic areas. Place2Vec \citep{place2vec} learns vectors for place categories, such as auto-repair shops and gas stations, from the augmented spatial context of surrounding points of interest. Its output represents a category, rather than an individual establishment or an arbitrary coordinate. hex2vec \citep{hex2vec} learns skip-gram embeddings of H3 hexagons from aggregated OpenStreetMap tags.

Related hybrid approaches combine place information with imagery. Urban2Vec \citep{urban2vec} learns neighborhood embeddings from POI information and street-view images.

Map attributes can also be used directly, without learning an embedding. The srai CountEmbedder \citep{srai} counts OpenStreetMap tags within supplied polygons. Area-based embeddings and direct map descriptors summarize predefined hexagons or polygons, with locations within each area sharing its representation. CartoVec instead learns a continuous coordinate-to-vector function, so new coordinates can be queried and their vectors aggregated without constructing a map descriptor for each new unit.

\textbf{Evaluation methodology.} Frozen-encoder evaluations assess reusable features by fitting a downstream prediction model without updating the encoder, as in SatCLIP and GeoCLIP \citep{satclip,geoclip}. TorchSpatial \citep{torchspatial} provides implementations, pretrained models and benchmarks for location encoders. We use its implementations and checkpoints for the Space2Vec and Sphere2Vec baselines, while defining our own downstream tasks and protocol.

The geographic structure of the split also matters. Nearby observations can share environmental conditions and similar outcomes. \citet{spatialcv} show that ignoring dependence between observations can underestimate prediction error and recommend blocked cross-validation for dependent data. \citet{spatialvalidation} find that apparent performance in forest-biomass mapping declines sharply under spatial validation, illustrating how random splits can reward reliance on geographic proximity. Following this rationale, we use spatial block cross-validation to evaluate encoders on held-out tiles within the study area. All records from a tile stay in the same fold, reducing the mixing of nearby observations across training and test sets.

\section{Method}\label{method}

\subsection{Problem setup and encoder}\label{problem-setup-and-encoder}

We learn a deterministic coordinate encoder \(f_\theta : \mathbb{R}^{2} \to \mathbb{R}^{64}\), mapping latitude and longitude in degrees to a 64-dimensional, unit-normalized embedding. The model has three parts: a fixed geometric transform of the coordinate, a multilayer perceptron, and, during training only, a decoder with prediction heads. Figure~\ref{fig:1} summarizes the architecture and layer widths.

\textbf{Coordinate representation.} The encoder accepts latitude and longitude in degrees and converts them to radians by multiplying by \(\pi/180\). The resulting latitude \(\varphi\) and longitude \(\lambda\) are mapped to the unit vector \(u = (\cos \varphi \cos \lambda, \cos \varphi \sin \lambda, \sin \varphi)\) in an Earth-centered, Earth-fixed frame, using a spherical Earth approximation. This representation avoids the coordinate discontinuity at the antimeridian and the longitude singularity at the poles, following the geometric motivation of spherical location encoders \citep{sphere2vec}.

\textbf{Positional features.} The unit vector is expanded into deterministic Fourier features \(\{\sin(\pi \cdot 2^{k}\cdot u), \cos(\pi \cdot 2^{k}\cdot u)\}\) for \(k = 0 \ldots K-1\), applied componentwise to \(u\). These sinusoids supply coordinate variation at multiple scales: low-frequency bands change gradually, while high-frequency bands distinguish more closely spaced locations. Providing this basis makes fine spatial variation easier for the network to represent than raw coordinates alone \citep{fourier}. We use \(K = 20\), giving \(3 \times  2 \times  20 = 120\) Fourier features, and concatenate the original three components of \(u\) to form the 123-dimensional network input.

\textbf{Encoder.} The positional features pass through a multilayer perceptron with residual blocks and a linear projection to \(\mathbb{R}^{64}\). The result is L2-normalized; the decoder consumes this normalized vector during training, and the same vector is served at inference. Pre-normalization activations are used for regularization.

\textbf{Decoder and heads.} During training, a shared decoder consumes the embedding and feeds separate output heads for the target groups described below. Encoder and decoder are trained end to end. The decoder is discarded after training; only the encoder is needed at inference.

\subsection{Training targets from Overture Maps}\label{training-targets-from-overture-maps}

The encoder is supervised by predicting map-derived attributes at and around each sampled coordinate. The decoder must recover these attributes from the embedding alone, encouraging the representation to retain information about land characteristics, transportation and nearby places.

The targets comprise binary indicators and counts. Base attributes are obtained by intersecting features with the sampled point; Transportation targets use proximity thresholds; Places counts use H3 neighborhoods. For Land and Water, Overture uses subtype for the broader grouping and class for the more detailed feature type; both are encoded as separate binary indicators. Each group has a separate prediction head, with primary and alternate Places categories modeled separately:

\begin{itemize}
\item \textbf{Land.} Binary indicators of Base land subtypes, such as land, forest and physical, and classes, such as island, beach and cliff, intersecting the coordinate. For example, a cliff has subtype physical and class cliff. These describe physical land features rather than cadastral parcels.
\item \textbf{Water.} Binary indicators of Base water subtypes, such as ocean, lake and human-made, and classes, such as bay, canal and reservoir, intersecting the coordinate. Together, Land and Water characterize the location's physical setting.
\item \textbf{Land cover.} Binary indicators of the physical surface classes of Base land-cover polygons containing the coordinate, such as forest, cropland and barren ground.
\item \textbf{Land use.} Binary indicators of the human uses assigned to Base land-use polygons containing the coordinate, such as residential, commercial and recreational use. Land cover and land use distinguish physical surface characteristics from how an area is used.
\item \textbf{Transportation segment type.} Binary indicators of segment subtype and class combinations within fixed proximity thresholds. Segments represent transportation paths, including roads, footpaths, railways and water routes. These targets describe the types of network infrastructure close to a location.
\item \textbf{Transportation connectors.} Two targets count network connector points within fixed proximity thresholds. Connectors include points where segments join and segment endpoints, so the counts provide local network context rather than counts of street intersections alone.
\item \textbf{Places.} Counts of points of interest by category at three spatial scales. We use H3, a hierarchical hexagonal index \citep{h3}, and count places in the containing cell and all cells within three grid steps at resolutions 9, 11 and 13. The containing cell and three surrounding rings give \texttt{1\ +\ 6\ +\ 12\ +\ 18\ =\ 37} cells. A coordinate can lie near the boundary of its containing hexagon. Including three surrounding rings captures nearby places across that boundary and forms a roughly circular neighborhood. The approximate radius scales are 1~km, 150~m and 20~m, respectively, retaining broader area context alongside local place composition. We count primary and alternate categories separately. To reduce sparsity, we sum raw category counts into Overture's coarser Basic Categories within each scale, increasing observations per target without mixing neighborhood sizes.
\item \textbf{Base-label availability.} One binary indicator is defined independently for each of Land, Water, Land cover and Land use. It equals one when that group's category labels are all unavailable at the coordinate, and zero otherwise. For example, no Water labels may be available at a point on land, while its Land cover labels are available. The indicators record these cases separately; unavailable labels do not necessarily indicate incomplete mapping.
\end{itemize}

After aggregation, we retain targets with at least five positive observations (binary indicators) or non-zero observations (counts) in a subset of about one million training coordinates. The criterion is applied to every head and removes 7.2\% of targets. Table~\ref{tab:training-targets} shows the counts before aggregation, after aggregation and after this occurrence filter.

\FloatBarrier
\begin{table}[htbp]
\centering\small\setlength{\tabcolsep}{3pt}
\caption{\textbf{Training-target heads.} For the Places heads, ``raw'' is the number of per-category columns before category grouping, ``aggregated'' is the number after Basic-Category grouping within scale, and ``retained'' is the number satisfying the occurrence filter. Other heads are not aggregated.}
\label{tab:training-targets}
\begin{tabular}{@{}
  >{\raggedright\arraybackslash}p{0.28\linewidth}
  >{\raggedright\arraybackslash}p{0.17\linewidth}lrrr@{}}
\toprule
Group & \shortstack[l]{Overture\\theme} & Type & \shortstack[r]{Raw\\Columns}
& \shortstack[r]{Aggregated\\Targets} & \shortstack[r]{Retained\\Targets} \\
\midrule
Land & Base & binary & 57 & 57 & 44 \\
Water & Base & binary & 49 & 49 & 32 \\
Land cover & Base & binary & 11 & 11 & 7 \\
Land use & Base & binary & 135 & 135 & 116 \\
Transportation segment type & Transportation & binary & 52 & 52 & 52 \\
Transportation connectors & Transportation & count & 2 & 2 & 2 \\
Places (primary category) & Places & count & 5,625 & 1,002 & 947 \\
Places (alternate category) & Places & count & 5,481 & 992 & 933 \\
Base-label availability & Base & binary & 4 & 4 & 4 \\
\bottomrule
\end{tabular}
\end{table}

Aggregation reduces the 11,106 raw Places columns to 1,994 targets, approximately a 5.6-fold reduction, while preserving the three spatial scales. Before filtering there are 308 binary and 1,996 count targets in total; training retains 255 binary and 1,882 count targets. Most counts remain zero at any given coordinate, motivating the count likelihood below.

\subsection{Training objective}\label{training-objective}

Binary heads use masked binary cross-entropy, with positive-class weights estimated from the training data to account for imbalance. Count heads use a zero-inflated negative binomial (ZINB) likelihood \citep{zinb}, combining an additional probability mass at zero with an overdispersed count distribution. This accommodates both the many absent categories and variation in non-zero counts. For each count target, the decoder predicts the zero-inflation probability, the mean of the negative binomial component and its dispersion parameter from the embedding. All three vary with the input.

An observed zero is a valid target: it means that a category is absent or that its count is zero. An unavailable label is different. For example, if no Land use category is recorded at a coordinate, its unavailable category indicators are excluded from the loss using an observation mask, while the Land use availability indicator is still predicted. Each head's loss is averaged over its observed entries. Head averages are then weighted by their retained target counts within the binary and count families to form \(L_{\mathrm{binary}}\) and \(L_{\mathrm{count}}\). These two loss families are distinct from the thematic target groups in Table~\ref{tab:training-targets}.

The complete objective is

\begin{equation}\label{eq:objective}
L = L_{\mathrm{binary}} + L_{\mathrm{count}} + 0.1L_{\mathrm{jitter}}
    + 0.1L_{\mathrm{var}} + 0.05L_{\mathrm{cov}} + L_{\mathrm{center}}.
\end{equation}

\(L_{\mathrm{jitter}}\) encourages nearby coordinates to have similar embeddings. For a batch of \(N\) coordinates \(x_{i}\), let \(\tilde{x}_{i}\) be the result of adding independent uniform noise in \([-0.01^\circ, 0.01^\circ]\) to each latitude and longitude. Since the embeddings have unit norm, their inner product is their cosine similarity:

\begin{equation}\label{eq:jitter}
L_{\mathrm{jitter}} = \frac{1}{N}\sum_{i=1}^{N}
\left[1-\left\langle f_\theta(x_i),f_\theta(\tilde{x}_i)\right\rangle\right].
\end{equation}

The remaining penalties act on the pre-normalization activations \(z_{i}\). Let \(D = 64\), \(\bar{z}\) be their batch mean and \(C\) their sample covariance matrix. We define

\begin{equation}\label{eq:regularization}
\begin{aligned}
s_d &= \sqrt{\frac{1}{N}\sum_{i=1}^{N}(z_{id}-\bar z_d)^2+10^{-6}},\\
L_{\mathrm{var}} &= \frac{1}{D}\sum_{d=1}^{D}\max(0,1-s_d),\\
L_{\mathrm{cov}} &= \frac{1}{D}\sum_{d\ne d'}C_{dd'}^2,\\
L_{\mathrm{center}} &= \frac{1}{D}\lVert\bar z\rVert_2^2.
\end{aligned}
\end{equation}

These terms discourage low-variance dimensions, correlated dimensions and a common offset that can concentrate normalized embeddings around one direction. The variance and covariance penalties draw on the anti-collapse and redundancy-reduction principles of VICReg \citep{vicreg} and Barlow Twins \citep{barlow}.

The encoder and decoder are trained with Adam, weight decay and gradient clipping. Evaluation uses exponential-moving-average encoder weights.

No downstream evaluation labels enter this objective. The experiment tests whether compressing map attributes into a fixed-width coordinate function produces useful features for tasks not used in training.

\subsection{Data sampling and region}\label{data-sampling-and-region}

The training data contain approximately 30.7 million sampled coordinates, with targets derived from a single Overture Maps snapshot. The study area covers the US states of New York, New Jersey, Pennsylvania, Delaware, Maryland, Connecticut, Rhode Island, Massachusetts, Vermont and New Hampshire, together with the District of Columbia. We refer to this study area as the US Northeast; Maine is not included.

\begin{figure}[htbp]\centering
\includegraphics[width=\linewidth,height=0.65\textheight,keepaspectratio]{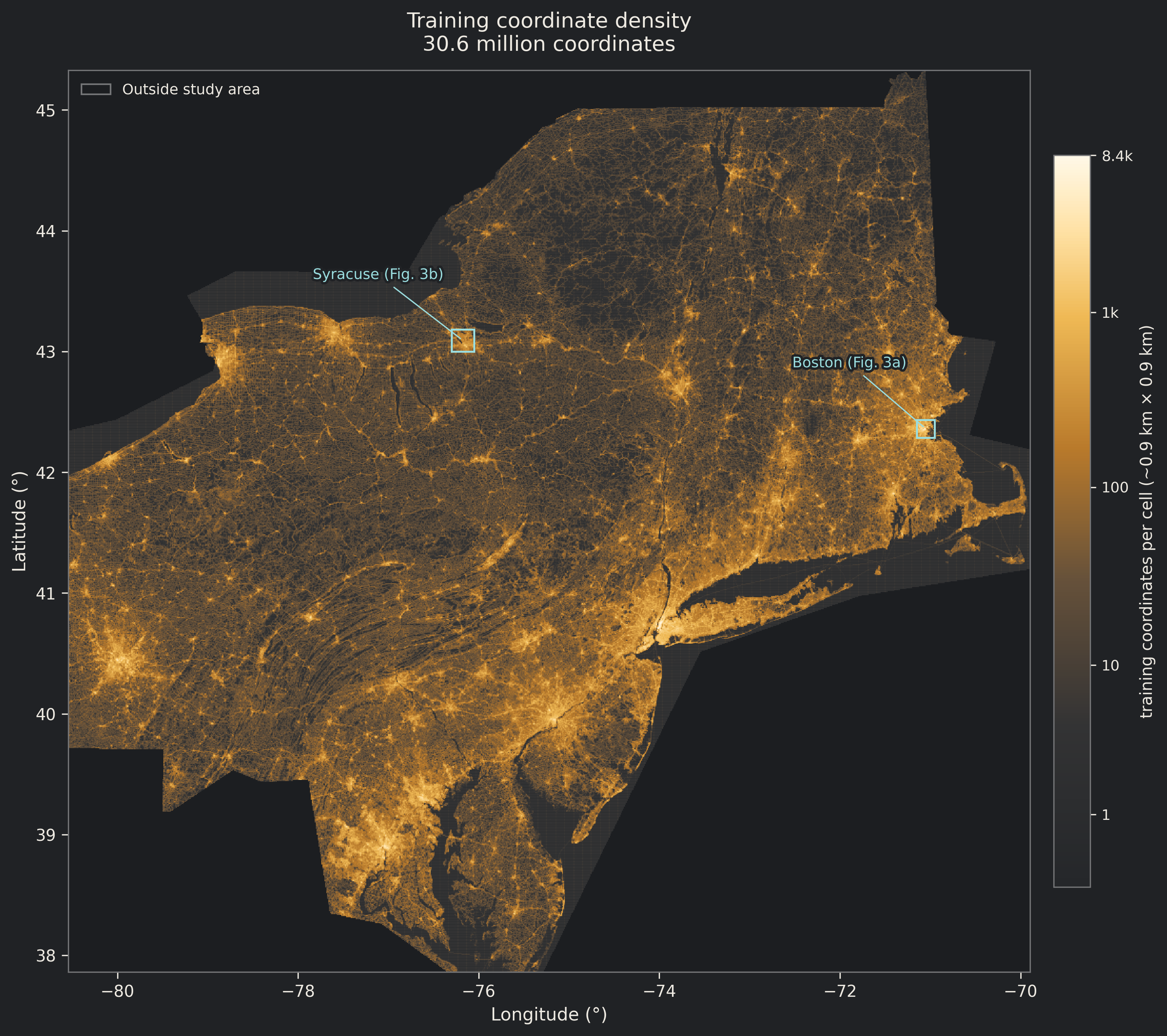}
\caption{Spatial distribution of training coordinates across the US Northeast. For display, coordinates are counted in raster cells approximately 0.9~km \(\times\) 0.9~km; these cells summarize sampling density. The color scale is logarithmic with a mild gamma stretch. Brighter amber-to-white colors indicate denser sampling; dark gray marks areas outside the study mask. Dense urban cores, rural road corridors and the background grid reveal the adaptive, feature-driven sampling. The outlined windows locate the Boston and Syracuse views in Figure~\ref{fig:3}(\subref{fig:3a}) and (\subref{fig:3b}), respectively.}
\label{fig:2}
\end{figure}

\begin{figure}[htbp]\centering
\captionsetup[subfigure]{position=top,skip=3pt}
\begin{subfigure}[t]{0.49\linewidth}
\centering
\caption{Boston}\label{fig:3a}
\includegraphics[width=\linewidth]{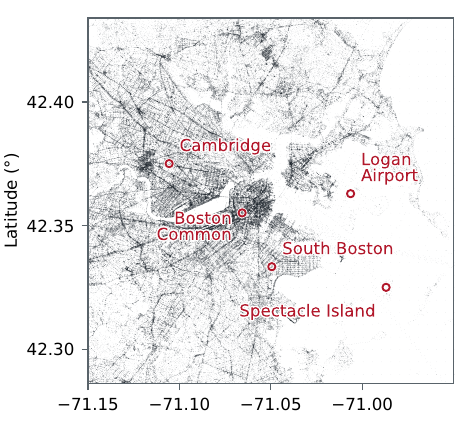}
\end{subfigure}\hfill
\begin{subfigure}[t]{0.49\linewidth}
\centering
\caption{Syracuse}\label{fig:3b}
\includegraphics[width=\linewidth]{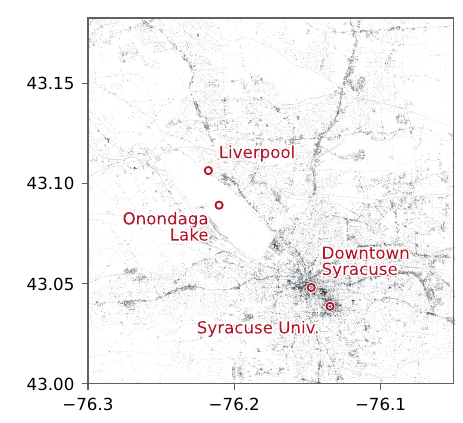}
\end{subfigure}
\par{\scriptsize Longitude ($^\circ$)\par}
\caption{Sampled training coordinates at local scale. (a) Boston, its harbor and nearby islands: a 16.5 \(\times\) 16.5~km window with 230,966 coordinates (853 per km\(^2\)). (b) Syracuse and Onondaga Lake: a 20.3 \(\times\) 20.3~km window with 101,586 coordinates (246 per km\(^2\)). Road-aligned samples, building/place clusters and a background lattice show how map features guide sampling. Syracuse's lower sampling density makes roads and the background grid more evident outside the city core; lake coverage is sparse. The panels contrast a metropolitan center with a less densely built area and have different geographic extents. Red circles identify landmarks for orientation.}
\label{fig:3}
\end{figure}

We first select coordinates using a background grid and feature-derived samples, then compute the map targets at each selected point. The grid uses 0.005\(^\circ\) spacing, approximately 555~m north--south and 390--440~m east--west across the study area. Denser samples are drawn from transportation features, buildings, places and Base features. This concentrates training coordinates in the built environment while retaining a sparse background over land. The resulting distribution is shown in Figure~\ref{fig:2}, with local views in Figure~\ref{fig:3}. Sampling density is not uniform, and water and other feature-poor areas are less represented. Each coordinate has the 2,137 retained targets described above, with unavailable entries masked during training.

\section{Experimental Design}\label{experimental-design}

We evaluate the frozen encoder alongside a set of publicly available baselines under a single, shared protocol: the same prediction units, spatial folds and validation-based rule for selecting the downstream model. Coordinate encoders also share the points used to represent each unit.

\subsection{Baselines}\label{baselines}

We compare CartoVec with nine baselines spanning three approaches:

\textbf{Coordinates and geometric features.}
\begin{itemize}
\item \textbf{Raw latitude/longitude baseline:} the coordinate pair scaled as \texttt{(lat\ /\ 90,\ lon\ /\ 180)}.
\item \textbf{RFF-256:} 256 random Fourier features \citep{rff} of the same rescaled coordinates, with bandwidth set by the median-pairwise-distance heuristic.
\end{itemize}

\textbf{Pretrained coordinate encoders.} All use frozen public checkpoints. For Space2Vec and Sphere2Vec, TorchSpatial \citep{torchspatial} provides variants pretrained for geographically informed image classification on Functional Map of the World (FMoW) satellite images and Yahoo Flickr Creative Commons (YFCC) photographs.
\begin{itemize}
\item \textbf{Space2Vec-grid FMoW} \citep{space2vec}: multi-scale planar grid features with FMoW pretraining.
\item \textbf{Space2Vec-grid YFCC:} the same positional encoding with YFCC pretraining.
\item \textbf{Sphere2Vec-sphereC+ FMoW} \citep{sphere2vec}: sphere-aware positional features with FMoW pretraining.
\item \textbf{Sphere2Vec-sphereC+ YFCC:} the same positional encoding with YFCC pretraining.
\item \textbf{SatCLIP-256} \citep{satclip}: ResNet50-L40. We select the finer \(L=40\) configuration over \(L=10\) following the authors' regional interpolation results.
\item \textbf{GeoCLIP-512} \citep{geoclip}: the released location encoder.
\end{itemize}

\textbf{Direct map counts.}
\begin{itemize}
\item \textbf{CountEmbedder} \citep{srai}: the srai descriptor counting OpenStreetMap tags within supplied polygons, retaining all 725 count columns in our configuration.
\end{itemize}

Eight baselines take coordinates as input, sharing CartoVec's query interface; the ninth, CountEmbedder, constructs features from map records at each prediction unit. Encoder settings remain fixed; downstream probe architectures are selected on validation data.

\subsection{Benchmark and tasks}\label{benchmark-and-tasks}

We evaluate on three task families, all inside the study region. The main benchmark uses the 2018--2022 US Census American Community Survey 5-year estimates \citep{acs} at block-group resolution, across six demographic and socioeconomic targets: population density, median household income, median home value, median gross rent, educational attainment (percentage of residents aged 25 or older with a bachelor's degree or higher), and median age. The study region yields 48,087 block groups. Population density is computed from ACS population and TIGER/Line 2022 land area \citep{tiger}, excluding water area. Density, income, home value and rent use natural logarithms; education and age remain on their original scales.

We also evaluate Airbnb nightly price prediction at listing points and Philadelphia crime-count prediction over H3 resolution-10 cells, using tasks derived from OBSR \citep{obsr}. Both use \texttt{log1p} targets and no clipping or winsorization. The Philadelphia unit set includes zero-crime cells rather than conditioning on cells where an incident occurred, so the task measures both separating zero from non-zero cells and predicting intensity. All test \(R^2\) scores are computed on the target scale after any transformation.

\textbf{Representing prediction units.} For Census block groups and Philadelphia H3 cells, we average each coordinate encoder's outputs over points sampled inside the polygon. For Census block-group land area \(A\) in km\(^2\), the sample count is \(n(A) = \min(1024, \max(32, \lceil 32A \rceil))\). Census uses 8,009,841 points across its 48,087 units, approximately 167 per block group; Philadelphia cells use 32 points each. All coordinate encoders share the same cached points, sampled uniformly within each polygon in latitude-longitude space. The resulting vectors are arithmetic means and are not renormalized. Airbnb uses the listing coordinate directly. CountEmbedder instead counts OSM features intersecting each polygon, or a buffer of 1~km radius around each Airbnb listing.

\subsection{Probing and cross-validation}\label{probing-and-cross-validation}

Each frozen encoder embeds the evaluation coordinates. For each encoder and target, we fit downstream prediction models to the fixed embeddings. We call these models probes.

\textbf{Spatial splits.} We group prediction units into 0.05\(^\circ\) tiles for Census and 0.02\(^\circ\) tiles for Airbnb and Philadelphia crime. Units are assigned to tiles using Census-provided internal points (centroids or nearby interior points) for block groups, H3 cell centers for crime, and listing coordinates for Airbnb. Tiles are randomly ordered and assigned to five groups, called folds, balancing unit counts. Each fold contains tiles distributed across the study region. In each evaluation round, one fold is held out as the test set. The other four folds form the training pool, containing the units available for training and validation in that round. We reserve approximately 12.5\% of this pool for validation, also by whole tile, and use the rest for training. Training data fit the probe; validation data guide early stopping and architecture selection; test data measure predictive performance.

\begin{figure}[htbp]\centering
\includegraphics[width=\linewidth,height=0.65\textheight,keepaspectratio]{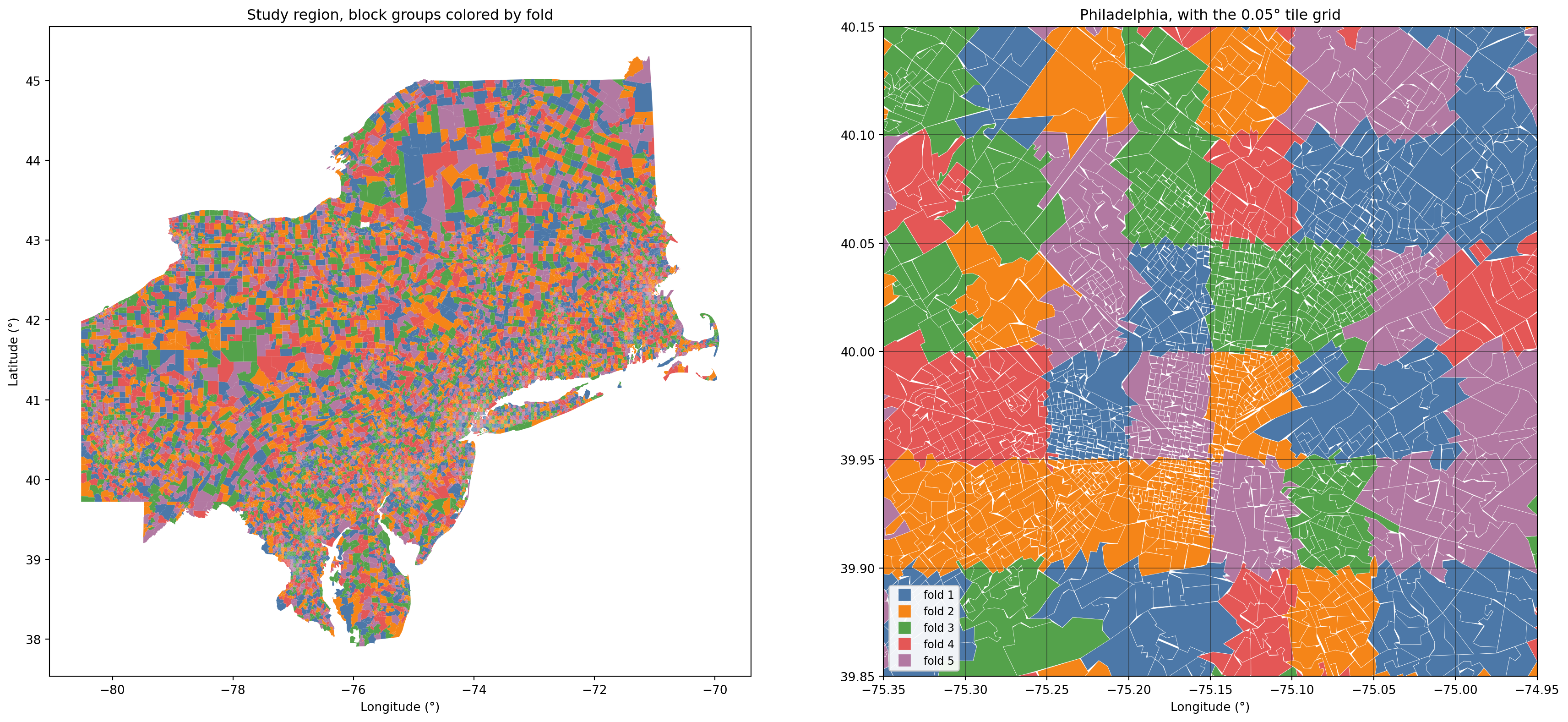}
\caption{Spatial block cross-validation for the Census tasks. Left: all 48,087 Census block groups in the evaluation region, colored by their assigned fold. Right: a closer view around Philadelphia, with the 0.05\(^\circ\) tile grid in dark lines and block-group boundaries in white. Each fold is held out for testing in turn; the remaining tiles are divided between training and validation. Block groups whose representative coordinates fall in the same tile share its fold, although their polygon boundaries can cross tile edges.}
\label{fig:4}
\end{figure}

We repeat this procedure until each fold has served as the test set once, producing five cross-validation splits. Each split contains approximately 70\% training, 10\% validation and 20\% test units. Tile grouping does not impose a geographic gap between training and test units. Figure~\ref{fig:4} shows the five Census folds across the study region and in a closer view around Philadelphia.

\textbf{Probe selection.} The probe's width and depth affect how much target-relevant information it can recover, and the suitable architecture need not be the same for every encoder. For each encoder and target, we use the five architectures in Table~\ref{tab:probe-architectures}, following the validation-based downstream tuning used by SatCLIP \citep{satclip}. All candidates use mean-squared-error loss, the Adam optimizer, learning-rate reduction and early stopping on validation loss. Hidden layers use LeakyReLU, batch normalization, dropout and an L2 kernel penalty. Probe inputs receive no additional standardization. The weights with the lowest validation loss are restored before testing.

\begin{table}[htbp]
\centering\small
\caption{Downstream probe architectures. Hidden-layer entries give successive widths; every probe has a linear regression output.}
\label{tab:probe-architectures}
\begin{tabular}{@{}ll@{}}
\toprule
Architecture & Hidden layers \\
\midrule
linear & none (a linear map on the embedding) \\
small & \(64 \rightarrow 32\) \\
base & \(256 \rightarrow 128\) \\
wide & \(1024 \rightarrow 512\) \\
geoclip-shape & 2048 (single layer) \\
\bottomrule
\end{tabular}
\end{table}

The \texttt{geoclip-shape} architecture matches the single 2,048-unit hidden layer used in GeoCLIP's frozen-feature classification evaluation \citep[Section~4.5]{geoclip}.

For each encoder and target, we select the architecture with the highest mean validation \(R^2\) across the five evaluation rounds and report that architecture's test scores on the same splits. This applies the same selection rule across encoders while allowing the prediction network to suit each representation.

\section{Results and findings}\label{results-and-findings}

\subsection{Predictive performance across tasks}\label{predictive-performance-across-tasks}

Table~\ref{tab:downstream} reports the final comparison on 48,087 Census block groups, 55,299 Airbnb listings and 24,786 Philadelphia H3 cells. 

\FloatBarrier
\begin{table}[htbp]
\centering\small\setlength{\tabcolsep}{2pt}
\caption{\textbf{Downstream prediction.} Mean test \(R^2\) over 5 spatial folds. Each method is reported at the probe architecture selected for it on validation (Section~\ref{probing-and-cross-validation}). Bold marks the highest mean in each column. The macro-average is the unweighted mean across the eight tasks. \(R^2\) for transformed targets is measured on the transformed scale (Section~\ref{benchmark-and-tasks}).}
\label{tab:downstream}
\begin{tabular}{@{}l*{9}{r}@{}}
\toprule
Method & \shortstack[r]{Pop.\\density} & \shortstack[r]{Educ.\\attainment}
& \shortstack[r]{Median\\age} & \shortstack[r]{Median\\household\\income}
& \shortstack[r]{Median\\home\\value} & \shortstack[r]{Median\\gross\\rent}
& \shortstack[r]{Airbnb\\price} & \shortstack[r]{Phila.\\crime}
& \shortstack[r]{Macro-\\average} \\
\midrule
\textbf{CartoVec-64} & \textbf{0.883} & \textbf{0.492} & \textbf{0.158} & \textbf{0.429} & \textbf{0.732} & 0.434 & \textbf{0.234} & 0.451 & \textbf{0.477} \\
\addlinespace[3pt]
Raw lat/lon & 0.441 & 0.095 & 0.035 & 0.063 & 0.488 & 0.327 & -0.004 & -0.104 & 0.168 \\
RFF-256 & 0.631 & 0.188 & 0.057 & 0.152 & 0.610 & 0.390 & 0.219 & 0.343 & 0.324 \\
CountEmb-725 & 0.836 & 0.318 & 0.142 & 0.292 & 0.429 & 0.256 & 0.209 & \textbf{0.594} & 0.384 \\
\shortstack[l]{Space2Vec FMoW} & 0.625 & 0.146 & 0.060 & 0.139 & 0.586 & 0.382 & 0.187 & 0.056 & 0.273 \\
\shortstack[l]{Space2Vec YFCC} & 0.686 & 0.347 & 0.095 & 0.219 & 0.657 & 0.413 & 0.205 & 0.143 & 0.346 \\
\shortstack[l]{Sphere2Vec FMoW} & 0.691 & 0.362 & 0.088 & 0.185 & 0.658 & 0.406 & 0.209 & 0.258 & 0.357 \\
\shortstack[l]{Sphere2Vec YFCC} & 0.616 & 0.145 & 0.053 & 0.134 & 0.588 & 0.376 & 0.142 & -0.008 & 0.256 \\
SatCLIP-256 & 0.625 & 0.167 & 0.054 & 0.152 & 0.596 & 0.379 & 0.198 & 0.182 & 0.294 \\
GeoCLIP-512 & 0.787 & 0.478 & 0.140 & 0.390 & 0.729 & \textbf{0.450} & 0.231 & 0.386 & 0.449 \\
\bottomrule
\end{tabular}
\end{table}

\textbf{Map supervision supports task reuse.} CartoVec has the highest mean \(R^2\) on five of six Census targets and the highest unweighted average across the eight tasks (Table~\ref{tab:downstream}). GeoCLIP has the highest mean on gross rent and is the strongest image-grounded encoder in this suite.

\textbf{Fold consistency.} CartoVec beats every alternative on all five folds for population density and household income. For median age, it wins 4/5 folds against CountEmb, the highest-scoring alternative, and 5/5 against every other method. Comparisons with GeoCLIP, the highest-scoring alternative for home value, education and Airbnb price, are mixed (3/5 each); we describe these as \emph{at par}. GeoCLIP leads on rent (4/5), while CountEmb leads on crime (5/5). A consistent lead means at least four fold wins; \emph{at par} describes 2-3 wins.

\textbf{Direct counts can remain preferable.} CountEmb has the highest crime score. It retains local OSM counts within prediction cells, whereas CartoVec averages learned vectors over those cells. CountEmb also has 725 inputs versus 64 and uses a different map source. These differences prevent attributing the gap solely to retaining counts.

\textbf{Both geometry and pretraining provenance matter.} Untrained RFF features recover substantial signal once paired with a nonlinear probe. Space2Vec and Sphere2Vec differ between FMoW and YFCC checkpoints, and SatCLIP is below GeoCLIP on every target. The ordering therefore concerns these pretrained configurations rather than supervision modalities in isolation.

\subsection{Sensitivity to probe architecture}\label{interpreting-the-encoder-comparison}

Across the eight tasks, CartoVec has the highest macro-average at each of the four nonlinear probe architectures, while GeoCLIP leads with a linear probe. The validation-selected architectures also differ: CartoVec and GeoCLIP select the base network on several Census targets, whereas RFF, Space2Vec and Sphere2Vec select the widest two-layer network on all six. These differences support selecting probes per encoder rather than assuming that one architecture suits every representation. Appendix~\ref{app:analyses} reports the complete comparison and checks the aggregate findings using fold medians and task ranks.

\subsection{Where the embedding helps, and where it does not}\label{where-the-embedding-helps-and-where-it-does-not}

CartoVec achieves its highest \(R^2\) on population density, an outcome closely related to the land use, transportation and place composition used as supervision. This is consistent with the embedding retaining built-environment context. Its performance on income, Airbnb prices and crime counts also shows utility for outcomes not included in its training targets. These predictive associations do not identify the contributions of individual map attributes or establish causal relationships.

Housing and listing prices give a more mixed comparison: CartoVec and GeoCLIP are at par on home value and Airbnb price under the fold-count convention, while GeoCLIP leads on gross rent. Map composition supplies context associated with prices, but does not directly measure property characteristics or market conditions. GeoCLIP's photographic pretraining \citep{geoclip} may provide additional cues, such as building appearance and streetscape characteristics, that our category and count targets do not record. Such cues could distinguish areas with similar amenities but different socioeconomic conditions, offering a plausible complement to map composition for housing prediction. Multimodal neighborhood representations combining street-view and place information also support socioeconomic prediction \citep{urban2vec}.

Median age illustrates a different distinction: CartoVec leads among the evaluated methods, yet its absolute \(R^2\) is only 0.158, leaving most variation unexplained in this setup.

\subsection{Limitations and responsible use}\label{limitations-and-responsible-use}

The evaluation covers a small, relatively map-rich part of the world. We restricted data preparation and training to the US Northeast to keep computation manageable. More computational resources could support broader coverage, denser coordinate sampling and larger networks, potentially improving spatial detail and downstream performance. The present results establish the usefulness of map-derived supervision at this scale; the benefits of further scaling remain to be quantified.

The map snapshot post-dates the 2018–2022 ACS estimates, the March 2023 Airbnb listings and the 2023 Philadelphia crime incidents. Urban development, business turnover and changes in mapping coverage may therefore introduce features unavailable during the target periods. Changes occurring after the snapshot are not reflected in the encoder. Adaptive sampling concentrates supervision in built areas, with sparser coverage over water and feature-poor land.

Coverage depends partly on volunteer mapping and data-provider activity. Uneven coverage can make the representation more informative for some communities than others. Although training uses map attributes rather than personal characteristics, location features can encode socioeconomic structure correlated with protected attributes. Predictive performance on aggregate outcomes does not establish suitability for individual decisions in credit, insurance or policing. Such applications need task-specific validation and fairness assessment; average benchmark scores alone do not establish comparable performance across communities.

\section{Conclusion}\label{conclusion}

CartoVec learns a reusable coordinate representation by predicting structured open map attributes, without image inputs. It is competitive with image-grounded encoders and has the highest average score across the evaluated tasks, although several pairwise differences are close and a direct map-count baseline leads on one of the eight tasks. These results establish open map data as a useful source of task-reusable coordinate embeddings. The compact inference interface is useful when prediction units change or new records must be embedded, while spatial scale, map coverage and task context still govern its suitability.

The open-weight CartoVec model is available at \href{https://huggingface.co/gursoyfurkan/CartoVec}{huggingface.co/gursoyfurkan/CartoVec}.

\printbibliography[title={References}]

@inproceedings{vicreg,
  author    = {Bardes, A. and Ponce, J. and LeCun, Y.},
  title     = {{VICReg}: Variance-invariance-covariance regularization for self-supervised learning},
  booktitle = {International Conference on Learning Representations},
  date      = {2022},
  url       = {https://arxiv.org/abs/2105.04906}
}

@inproceedings{srai,
  author    = {Gramacki, P. and Le{\'s}niara, K. and Raczycki, K. and Wo{\'z}niak, S. and Przymus, M. and Szyma{\'n}ski, P.},
  title     = {{SRAI}: Towards standardization of geospatial {AI}},
  booktitle = {Proceedings of the 6th ACM SIGSPATIAL International Workshop on AI for Geographic Knowledge Discovery},
  publisher = {Association for Computing Machinery},
  date      = {2023},
  doi       = {10.1145/3615886.3627740}
}

@report{zinb,
  author      = {Greene, W. H.},
  title       = {Accounting for excess zeros and sample selection in {Poisson} and negative binomial regression models},
  type        = {Working Paper},
  number      = {EC-94-10},
  institution = {New York University, Stern School of Business},
  date        = {1994},
  url         = {https://archive.nyu.edu/handle/2451/26263}
}

@article{osm-background,
  author       = {Haklay, M. and Weber, P.},
  title        = {{OpenStreetMap}: User-generated street maps},
  journaltitle = {IEEE Pervasive Computing},
  volume       = {7},
  number       = {4},
  pages        = {12--18},
  date         = {2008},
  doi          = {10.1109/MPRV.2008.80}
}

@article{satclip,
  author       = {Klemmer, K. and Rolf, E. and Robinson, C. and Mackey, L. and Ru{\ss}wurm, M.},
  title        = {{SatCLIP}: Global, general-purpose location embeddings with satellite imagery},
  journaltitle = {Proceedings of the AAAI Conference on Artificial Intelligence},
  volume       = {39},
  number       = {4},
  pages        = {4347--4355},
  date         = {2025},
  doi          = {10.1609/aaai.v39i4.32457}
}

@article{locationreview,
  author       = {Mai, G. and Janowicz, K. and Hu, Y. and Gao, S. and Yan, B. and Zhu, R. and Cai, L. and Lao, N.},
  title        = {A review of location encoding for {GeoAI}: Methods and applications},
  journaltitle = {International Journal of Geographical Information Science},
  volume       = {36},
  number       = {4},
  pages        = {639--673},
  date         = {2022},
  doi          = {10.1080/13658816.2021.2004602}
}

@inproceedings{space2vec,
  author    = {Mai, G. and Janowicz, K. and Yan, B. and Zhu, R. and Cai, L. and Lao, N.},
  title     = {Multi-scale representation learning for spatial feature distributions using grid cells},
  booktitle = {International Conference on Learning Representations},
  date      = {2020},
  url       = {https://arxiv.org/abs/2003.00824}
}

@inproceedings{csp,
  author    = {Mai, G. and Lao, N. and He, Y. and Song, J. and Ermon, S.},
  title     = {{CSP}: Self-supervised contrastive spatial pre-training for geospatial-visual representations},
  editor    = {Krause, A. and Brunskill, E. and Cho, K. and Engelhardt, B. and Sabato, S. and Scarlett, J.},
  booktitle = {Proceedings of the 40th International Conference on Machine Learning},
  volume    = {202},
  pages     = {23498--23515},
  publisher = {PMLR},
  date      = {2023},
  url       = {https://proceedings.mlr.press/v202/mai23a.html}
}

@article{sphere2vec,
  author       = {Mai, G. and Xuan, Y. and Zuo, W. and He, Y. and Song, J. and Ermon, S. and Janowicz, K. and Lao, N.},
  title        = {{Sphere2Vec}: A general-purpose location representation learning over a spherical surface for large-scale geospatial predictions},
  journaltitle = {ISPRS Journal of Photogrammetry and Remote Sensing},
  volume       = {202},
  pages        = {439--462},
  date         = {2023},
  doi          = {10.1016/j.isprsjprs.2023.06.016}
}

@inproceedings{obsr,
  author    = {Moska, J. and Furman, O. and Kozaczko, K. and Leszkiewicz, S. and Polczyk, J. and Gramacki, P. and Szyma{\'n}ski, P.},
  title     = {{OBSR}: Open benchmark for spatial representations},
  booktitle = {Proceedings of the 33rd ACM International Conference on Advances in Geographic Information Systems},
  pages     = {670--681},
  publisher = {Association for Computing Machinery},
  date      = {2025},
  doi       = {10.1145/3748636.3762783}
}

@online{overture,
  author     = {{Overture Maps Foundation}},
  title      = {{Overture Maps} (release 2025-12-17.0)},
  titleaddon = {Data set},
  date       = {2025-12-17},
  url        = {https://docs.overturemaps.org/blog/2025/12/17/release-notes/}
}

@article{spatialvalidation,
  author       = {Ploton, P. and Mortier, F. and R{\'e}jou-M{\'e}chain, M. and Barbier, N. and Picard, N. and Rossi, V. and Dormann, C. and Cornu, G. and Viennois, G. and Bayol, N. and Lyapustin, A. and Gourlet-Fleury, S. and P{\'e}lissier, R.},
  title        = {Spatial validation reveals poor predictive performance of large-scale ecological mapping models},
  journaltitle = {Nature Communications},
  volume       = {11},
  eid          = {4540},
  date         = {2020},
  doi          = {10.1038/s41467-020-18321-y}
}

@inproceedings{rff,
  author    = {Rahimi, A. and Recht, B.},
  title     = {Random features for large-scale kernel machines},
  booktitle = {Advances in Neural Information Processing Systems},
  volume    = {20},
  date      = {2007},
  url       = {https://papers.nips.cc/paper_files/paper/2007/hash/013a006f03dbc5392effeb8f18fda755-Abstract.html}
}

@article{spatialcv,
  author       = {Roberts, D. R. and Bahn, V. and Ciuti, S. and Boyce, M. S. and Elith, J. and Guillera-Arroita, G. and Hauenstein, S. and Lahoz-Monfort, J. J. and Schr{\"o}der, B. and Thuiller, W. and Warton, D. I. and Wintle, B. A. and Hartig, F. and Dormann, C. F.},
  title        = {Cross-validation strategies for data with temporal, spatial, hierarchical, or phylogenetic structure},
  journaltitle = {Ecography},
  volume       = {40},
  number       = {8},
  pages        = {913--929},
  date         = {2017},
  doi          = {10.1111/ecog.02881}
}

@inproceedings{fourier,
  author    = {Tancik, M. and Srinivasan, P. P. and Mildenhall, B. and Fridovich-Keil, S. and Raghavan, N. and Singhal, U. and Ramamoorthi, R. and Barron, J. T. and Ng, R.},
  title     = {Fourier features let networks learn high frequency functions in low dimensional domains},
  booktitle = {Advances in Neural Information Processing Systems},
  volume    = {33},
  date      = {2020},
  url       = {https://arxiv.org/abs/2006.10739}
}

@online{h3,
  author     = {{Uber}},
  title      = {{H3}: Hexagonal hierarchical geospatial indexing system},
  titleaddon = {Software documentation},
  url        = {https://h3geo.org/docs/},
  urldate    = {2026-10-09}
}

@online{tiger,
  author     = {{U.S. Census Bureau}},
  title      = {2022 {TIGER/Line} block-group shapefiles},
  titleaddon = {Data set},
  date       = {2022},
  url        = {https://www.census.gov/geographies/mapping-files/2022/geo/tiger-line-file.html}
}

@online{acs,
  author     = {{U.S. Census Bureau}},
  title      = {2018--2022 {American Community Survey} 5-year estimates},
  titleaddon = {Data set},
  date       = {2023},
  url        = {https://api.census.gov/data/2022/acs/acs5.html}
}

@inproceedings{geoclip,
  author    = {Vivanco Cepeda, V. and Nayak, G. K. and Shah, M.},
  title     = {{GeoCLIP}: {CLIP}-inspired alignment between locations and images for effective worldwide geo-localization},
  booktitle = {Advances in Neural Information Processing Systems},
  volume    = {36},
  date      = {2023},
  doi       = {10.52202/075280-0379}
}

@article{urban2vec,
  author       = {Wang, Z. and Li, H. and Rajagopal, R.},
  title        = {{Urban2Vec}: Incorporating street view imagery and {POIs} for multi-modal urban neighborhood embedding},
  journaltitle = {Proceedings of the AAAI Conference on Artificial Intelligence},
  volume       = {34},
  number       = {1},
  pages        = {1013--1020},
  date         = {2020},
  doi          = {10.1609/aaai.v34i01.5450}
}

@inproceedings{hex2vec,
  author    = {Wo{\'z}niak, S. and Szyma{\'n}ski, P.},
  title     = {{hex2vec}: Context-aware embedding {H3} hexagons with {OpenStreetMap} tags},
  booktitle = {Proceedings of the 4th ACM SIGSPATIAL International Workshop on AI for Geographic Knowledge Discovery},
  publisher = {Association for Computing Machinery},
  date      = {2021},
  doi       = {10.1145/3486635.3491076}
}

@inproceedings{torchspatial,
  author    = {Wu, N. and Cao, Q. and Wang, Z. and Liu, Z. and Qi, Y. and Zhang, J. and Ni, J. and Yao, X. and Ma, H. and Mu, L. and Ermon, S. and Ganu, T. and Nambi, A. and Lao, N. and Mai, G.},
  title     = {{TorchSpatial}: A location encoding framework and benchmark for spatial representation learning},
  booktitle = {Advances in Neural Information Processing Systems},
  volume    = {37},
  date      = {2024},
  doi       = {10.52202/079017-2588}
}

@inproceedings{place2vec,
  author    = {Yan, B. and Janowicz, K. and Mai, G. and Gao, S.},
  title     = {From {ITDL} to {Place2Vec}: Reasoning about place type similarity and relatedness by learning embeddings from augmented spatial contexts},
  booktitle = {Proceedings of the 25th ACM SIGSPATIAL International Conference on Advances in Geographic Information Systems},
  publisher = {Association for Computing Machinery},
  date      = {2017},
  doi       = {10.1145/3139958.3140054}
}

@inproceedings{gps2vec,
  author    = {Yin, Y. and Liu, Z. and Zhang, Y. and Wang, S. and Shah, R. R. and Zimmermann, R.},
  title     = {{GPS2Vec}: Towards generating worldwide {GPS} embeddings},
  booktitle = {Proceedings of the 27th ACM SIGSPATIAL International Conference on Advances in Geographic Information Systems},
  pages     = {416--419},
  publisher = {Association for Computing Machinery},
  date      = {2019},
  doi       = {10.1145/3347146.3359067}
}

@inproceedings{barlow,
  author    = {Zbontar, J. and Jing, L. and Misra, I. and LeCun, Y. and Deny, S.},
  title     = {{Barlow Twins}: Self-supervised learning via redundancy reduction},
  editor    = {Meila, M. and Zhang, T.},
  booktitle = {Proceedings of the 38th International Conference on Machine Learning},
  volume    = {139},
  pages     = {12310--12320},
  publisher = {PMLR},
  date      = {2021},
  url       = {https://proceedings.mlr.press/v139/zbontar21a.html}
}

\clearpage\appendix
\counterwithin{table}{section}
\renewcommand{\thetable}{\thesection\arabic{table}}

\section{Probe architecture and performance}\label{app:analyses}\label{app:probe-selections}

Does CartoVec's performance depend on the choice of prediction network? Table~\ref{tab:probe-capacity} reports the complete grid of ten methods, five architectures, eight tasks and five folds (2,000 fits), allowing comparisons at a fixed architecture as well as at the validation-selected architectures used in the main results.

\begin{table}[!ht]
\centering
\captionsetup{font=footnotesize,skip=4pt}
\caption{Complete probe comparison. Task entries are five-fold mean test \(R^2\); underlining marks validation-selected architectures, not the highest test scores. Macro mean and macro median average the eight tasks' fold means and fold medians, respectively. \enquote{2048} denotes the single-layer \texttt{geoclip-shape} probe (Table~\ref{tab:probe-architectures}).}
\label{tab:probe-capacity}\label{tab:probe-selections}
\fontsize{9}{9.7}\selectfont
\setlength{\tabcolsep}{2pt}
\renewcommand{\arraystretch}{1}
\begin{tabular}{@{}ll*{10}{r}@{}}
\toprule
Method & Probe & \shortstack[r]{Pop.\\density} & \shortstack[r]{Educ.\\attain.}
& \shortstack[r]{Median\\age} & \shortstack[r]{Household\\income}
& \shortstack[r]{Home\\value} & \shortstack[r]{Gross\\rent}
& \shortstack[r]{Airbnb\\price} & \shortstack[r]{Phila.\\crime}
& \shortstack[r]{Macro\\mean} & \shortstack[r]{Macro\\median} \\
\midrule
\textbf{CartoVec-64} & Linear & 0.793 & 0.178 & 0.104 & 0.249 & 0.509 & 0.334 & 0.206 & 0.413 & 0.348 & 0.345 \\
 & Small & 0.864 & 0.456 & \underline{0.158} & 0.388 & 0.687 & 0.417 & 0.233 & 0.445 & 0.456 & 0.454 \\
 & Base & 0.880 & \underline{0.492} & 0.149 & \underline{0.429} & 0.724 & \underline{0.434} & \underline{0.234} & 0.452 & 0.474 & 0.475 \\
 & Wide & \underline{0.883} & 0.480 & 0.116 & 0.415 & \underline{0.732} & 0.419 & 0.222 & 0.449 & 0.464 & 0.463 \\
 & 2048 & 0.876 & 0.481 & 0.121 & 0.379 & 0.718 & 0.355 & 0.208 & \underline{0.451} & 0.449 & 0.445 \\
\addlinespace[2pt]
Raw lat/lon & Linear & -0.013 & -0.012 & -0.003 & -0.122 & -0.095 & -0.128 & -0.049 & -0.038 & -0.057 & -0.053 \\
 & Small & 0.376 & 0.054 & 0.020 & 0.064 & 0.345 & 0.239 & \underline{-0.004} & -0.087 & 0.126 & 0.139 \\
 & Base & 0.424 & 0.067 & \underline{0.035} & \underline{0.063} & 0.399 & 0.280 & -0.069 & -0.199 & 0.125 & 0.143 \\
 & Wide & \underline{0.441} & \underline{0.095} & 0.023 & 0.021 & \underline{0.488} & \underline{0.327} & -0.061 & -0.519 & 0.102 & 0.111 \\
 & 2048 & 0.157 & 0.029 & -0.004 & 0.056 & 0.314 & 0.212 & -0.013 & \underline{-0.104} & 0.081 & 0.084 \\
\addlinespace[2pt]
RFF-256 & Linear & 0.327 & 0.030 & 0.009 & 0.084 & 0.445 & 0.321 & 0.163 & 0.211 & 0.199 & 0.199 \\
 & Small & 0.581 & 0.134 & 0.045 & 0.080 & 0.552 & 0.349 & \underline{0.219} & 0.366 & 0.291 & 0.292 \\
 & Base & 0.615 & 0.145 & 0.051 & 0.116 & 0.593 & 0.366 & 0.217 & 0.397 & 0.313 & 0.312 \\
 & Wide & \underline{0.631} & \underline{0.188} & \underline{0.057} & \underline{0.152} & \underline{0.610} & \underline{0.390} & 0.217 & \underline{0.343} & 0.324 & 0.325 \\
 & 2048 & 0.566 & 0.128 & 0.049 & 0.070 & 0.560 & 0.375 & 0.213 & 0.352 & 0.289 & 0.293 \\
\addlinespace[2pt]
CountEmb-725 & Linear & -3.488 & 0.071 & -1.196 & -24.182 & -30.912 & -26.770 & -0.006 & 0.528 & -10.744 & -1.984 \\
 & Small & 0.819 & 0.321 & \underline{0.142} & 0.265 & 0.389 & \underline{0.256} & 0.213 & \underline{0.594} & 0.375 & 0.372 \\
 & Base & 0.838 & 0.264 & 0.141 & 0.288 & 0.427 & 0.185 & 0.198 & 0.604 & 0.368 & 0.379 \\
 & Wide & \underline{0.836} & \underline{0.318} & 0.101 & \underline{0.292} & \underline{0.429} & 0.207 & \underline{0.209} & 0.592 & 0.373 & 0.373 \\
 & 2048 & 0.740 & 0.172 & -0.504 & 0.265 & 0.186 & 0.155 & 0.190 & 0.199 & 0.176 & 0.324 \\
\addlinespace[2pt]
Space2Vec FMoW & Linear & 0.479 & 0.055 & 0.027 & 0.108 & 0.555 & 0.369 & 0.049 & 0.040 & 0.210 & 0.208 \\
 & Small & 0.593 & 0.116 & 0.046 & 0.079 & 0.547 & 0.354 & 0.175 & 0.109 & 0.252 & 0.258 \\
 & Base & 0.609 & 0.130 & 0.054 & 0.098 & 0.552 & 0.375 & 0.188 & \underline{0.056} & 0.258 & 0.283 \\
 & Wide & \underline{0.625} & \underline{0.146} & \underline{0.060} & \underline{0.139} & \underline{0.586} & \underline{0.382} & \underline{0.187} & 0.071 & 0.274 & 0.273 \\
 & 2048 & 0.526 & 0.116 & 0.032 & 0.081 & 0.558 & 0.375 & 0.109 & -0.497 & 0.162 & 0.224 \\
\addlinespace[2pt]
Space2Vec YFCC & Linear & 0.541 & 0.091 & 0.038 & 0.130 & 0.556 & 0.365 & 0.150 & 0.226 & 0.262 & 0.265 \\
 & Small & 0.641 & 0.292 & 0.073 & 0.144 & 0.589 & 0.361 & 0.216 & 0.260 & 0.322 & 0.322 \\
 & Base & 0.664 & 0.330 & 0.082 & 0.179 & 0.619 & 0.387 & \underline{0.205} & \underline{0.143} & 0.326 & 0.329 \\
 & Wide & \underline{0.686} & \underline{0.347} & \underline{0.095} & \underline{0.219} & \underline{0.657} & \underline{0.413} & 0.192 & 0.179 & 0.348 & 0.340 \\
 & 2048 & 0.606 & 0.099 & 0.024 & 0.155 & 0.577 & 0.353 & 0.137 & 0.177 & 0.266 & 0.274 \\
\addlinespace[2pt]
Sphere2Vec FMoW & Linear & 0.525 & 0.092 & 0.040 & 0.092 & 0.518 & 0.343 & 0.117 & 0.171 & 0.237 & 0.239 \\
 & Small & 0.659 & 0.268 & 0.073 & 0.136 & 0.599 & 0.366 & 0.213 & 0.275 & 0.323 & 0.321 \\
 & Base & 0.678 & 0.333 & 0.085 & 0.192 & 0.647 & 0.390 & \underline{0.209} & \underline{0.258} & 0.349 & 0.347 \\
 & Wide & \underline{0.691} & \underline{0.362} & \underline{0.088} & \underline{0.185} & \underline{0.658} & \underline{0.406} & 0.205 & 0.257 & 0.357 & 0.352 \\
 & 2048 & 0.666 & 0.234 & 0.066 & 0.196 & 0.629 & 0.385 & 0.136 & 0.149 & 0.308 & 0.312 \\
\addlinespace[2pt]
Sphere2Vec YFCC & Linear & 0.488 & 0.080 & 0.039 & 0.101 & 0.543 & 0.365 & 0.077 & \underline{-0.008} & 0.211 & 0.211 \\
 & Small & 0.584 & 0.124 & 0.030 & 0.097 & 0.544 & 0.361 & \underline{0.142} & -0.621 & 0.158 & 0.241 \\
 & Base & 0.608 & 0.132 & 0.044 & 0.100 & 0.567 & 0.370 & 0.150 & -2.099 & -0.016 & 0.103 \\
 & Wide & \underline{0.616} & \underline{0.145} & \underline{0.053} & \underline{0.134} & \underline{0.588} & \underline{0.376} & 0.124 & -0.259 & 0.222 & 0.244 \\
 & 2048 & 0.486 & 0.097 & 0.033 & 0.084 & 0.554 & 0.362 & 0.105 & -0.442 & 0.160 & 0.233 \\
\addlinespace[2pt]
SatCLIP-256 & Linear & 0.428 & 0.065 & 0.037 & 0.096 & 0.523 & 0.357 & 0.020 & -0.050 & 0.184 & 0.190 \\
 & Small & 0.580 & 0.138 & 0.049 & 0.097 & 0.527 & 0.360 & 0.174 & -0.088 & 0.230 & 0.217 \\
 & Base & 0.612 & 0.154 & 0.053 & 0.129 & 0.592 & 0.371 & 0.194 & \underline{0.182} & 0.286 & 0.292 \\
 & Wide & \underline{0.625} & \underline{0.167} & \underline{0.054} & \underline{0.152} & \underline{0.596} & \underline{0.379} & \underline{0.198} & -0.064 & 0.263 & 0.270 \\
 & 2048 & 0.583 & 0.132 & 0.053 & 0.131 & 0.579 & 0.382 & 0.070 & 0.200 & 0.266 & 0.267 \\
\addlinespace[2pt]
GeoCLIP-512 & Linear & 0.707 & 0.261 & 0.095 & 0.288 & 0.625 & 0.405 & 0.224 & 0.295 & 0.363 & 0.362 \\
 & Small & 0.773 & 0.477 & 0.139 & 0.373 & 0.703 & 0.448 & \underline{0.231} & \underline{0.386} & 0.441 & 0.441 \\
 & Base & 0.790 & \underline{0.478} & \underline{0.140} & \underline{0.390} & 0.729 & \underline{0.450} & 0.235 & 0.393 & 0.451 & 0.457 \\
 & Wide & \underline{0.787} & 0.468 & 0.123 & 0.383 & \underline{0.729} & 0.436 & 0.212 & 0.393 & 0.441 & 0.444 \\
 & 2048 & 0.781 & 0.471 & 0.127 & 0.376 & 0.730 & 0.408 & 0.219 & 0.332 & 0.430 & 0.431 \\
\bottomrule
\end{tabular}
\end{table}

\small

CartoVec's aggregate lead persists across the four common nonlinear probe architectures, under both mean and median summaries. GeoCLIP leads with Linear, however, so a linear-only evaluation would give a different comparison. RFF also benefits substantially from a nonlinear probe, rising from a macro-average of 0.199 with Linear to 0.324 with Wide. These results show why the prediction model is part of the evaluation: its ability to use the fixed features can change the ordering of encoders.

The benefit of nonlinearity is task-dependent. CartoVec's \(R^2\) on educational attainment rises from 0.178 with Linear to 0.492 with Base, whereas its population-density \(R^2\) is already 0.793 with Linear. The representation therefore does not make every outcome equally accessible to a linear model. Some pairwise differences nevertheless persist across all architectures: CartoVec exceeds GeoCLIP on every population-density fold, while GeoCLIP leads on at least four gross-rent folds. Probe choice changes the size of these differences without reversing their direction.

Increasing capacity does not consistently improve performance. CartoVec's highest fixed-architecture average is at Base, and CountEmb's is at Small, rather than at 2048. Fold behavior also matters: CountEmb with 2048 has positive median-age test scores on three folds but negative scores on two, giving a mean of \(-0.504\) and a median of 0.065. Reporting both summaries makes this uneven performance across folds visible. A shared hidden width also does not equalize parameter counts, because embedding dimensions differ.

At the validation-selected architectures, CartoVec also leads by average task rank (1.25 versus 2.25 for GeoCLIP among ten methods; lower is better) and by the average of task-fold medians (0.473 versus 0.452). Its macro-average remains highest when any single task is omitted, so the aggregate lead is not dependent on one favorable task. Rank and score summaries are complementary: with 2048, GeoCLIP has the better average rank, but CartoVec has the higher macro-average. Ranks record ordering, whereas \(R^2\) averages also reflect the sizes of performance differences.

\end{document}